\documentclass[10pt,twocolumn,letterpaper]{article}

\usepackage[pagenumbers]{cvpr} 

\usepackage[accsupp]{axessibility}  

\usepackage{graphicx}
\usepackage{amsmath}
\usepackage{amssymb}
\usepackage{booktabs}
\usepackage{pifont}
\newcommand{\cmark}{\ding{51}}%
\newcommand{\xmark}{\ding{55}}%

\usepackage{xcolor}
\usepackage{color}
\usepackage{array}
\usepackage{booktabs}
\usepackage{makecell}
\usepackage{multirow}
\usepackage[flushleft]{threeparttable}
\def\x{$\times$}
\newcolumntype{x}[1]{>{\centering\arraybackslash}p{#1pt}}
\newlength\savewidth\newcommand\shline{\noalign{\global\savewidth\arrayrulewidth\global\arrayrulewidth 1pt}\hline\noalign{\global\arrayrulewidth\savewidth}}
\newcommand{\tablestyle}[2]{\setlength{\tabcolsep}{#1}\renewcommand{\arraystretch}{#2}\centering\footnotesize}

\newcommand{\red}{\textcolor{red}}

\newcommand{\zzrevise}[2]{{\color{black} #2}}

\usepackage[pagebackref,breaklinks,colorlinks]{hyperref}

\usepackage[capitalize]{cleveref}
\crefname{section}{Sec.}{Secs.}
\Crefname{section}{Section}{Sections}
\Crefname{table}{Table}{Tables}
\crefname{table}{Tab.}{Tabs.}

\newcommand{\residual}[2]{
  $\left[
    \begin{array}{cc}
      3 \times 3 \times 3, #1 \\
      3 \times 3 \times 3, #1
    \end{array}
  \right]
  \times #2$
}

\def\cvprPaperID{****} 
\def\confName{CVPR}
\def\confYear{2022}

\begin{document}


\title{Learning from Temporal Gradient for Semi-supervised  Action Recognition}

\author{
Junfei Xiao${^{1}}$
\qquad
Longlong Jing${^{2}}$
\qquad
Lin Zhang${^{3}}$
\qquad
Ju He${^{1}}$
\qquad
Qi She${^{4}}$

\\
Zongwei Zhou${^{1}}$

\qquad
Alan Yuille${^{1}}$
\qquad
Yingwei Li${^{1}}$
\\
\\$^{1}$Johns Hopkins University \qquad ~$^{2}$The City University of New York \\ ~$^{3}$Carnegie Mellon University \qquad ~$^{4}$ByteDance
}
\maketitle
\begin{abstract}

Semi-supervised video action recognition tends to enable deep neural networks to achieve remarkable performance even with very limited labeled data. However, existing methods are mainly transferred from current image-based methods (\eg, FixMatch). Without specifically utilizing the temporal dynamics and inherent multimodal attributes, their results could be suboptimal. To better leverage the encoded temporal information in videos, we introduce temporal gradient as an additional modality for more attentive feature extraction in this paper. To be specific, our method explicitly distills the fine-grained motion representations from temporal gradient (TG) and imposes consistency across different modalities (\ie, RGB and TG). The performance of semi-supervised action recognition is significantly improved without additional computation or parameters during inference.  Our method achieves the state-of-the-art performance on three video action recognition benchmarks (\ie, Kinetics-400, UCF-101, and HMDB-51) under several typical semi-supervised settings (\ie, different ratios of labeled data). Code 
is made available at \href{https://github.com/lambert-x/video-semisup}{https://github.com/lambert-x/video-semisup}.

\vspace{-5mm}
\end{abstract}

\begin{figure}[t!]
	\centering
	\renewcommand{\tabcolsep}{0pt}
	\includegraphics[width=1\columnwidth]{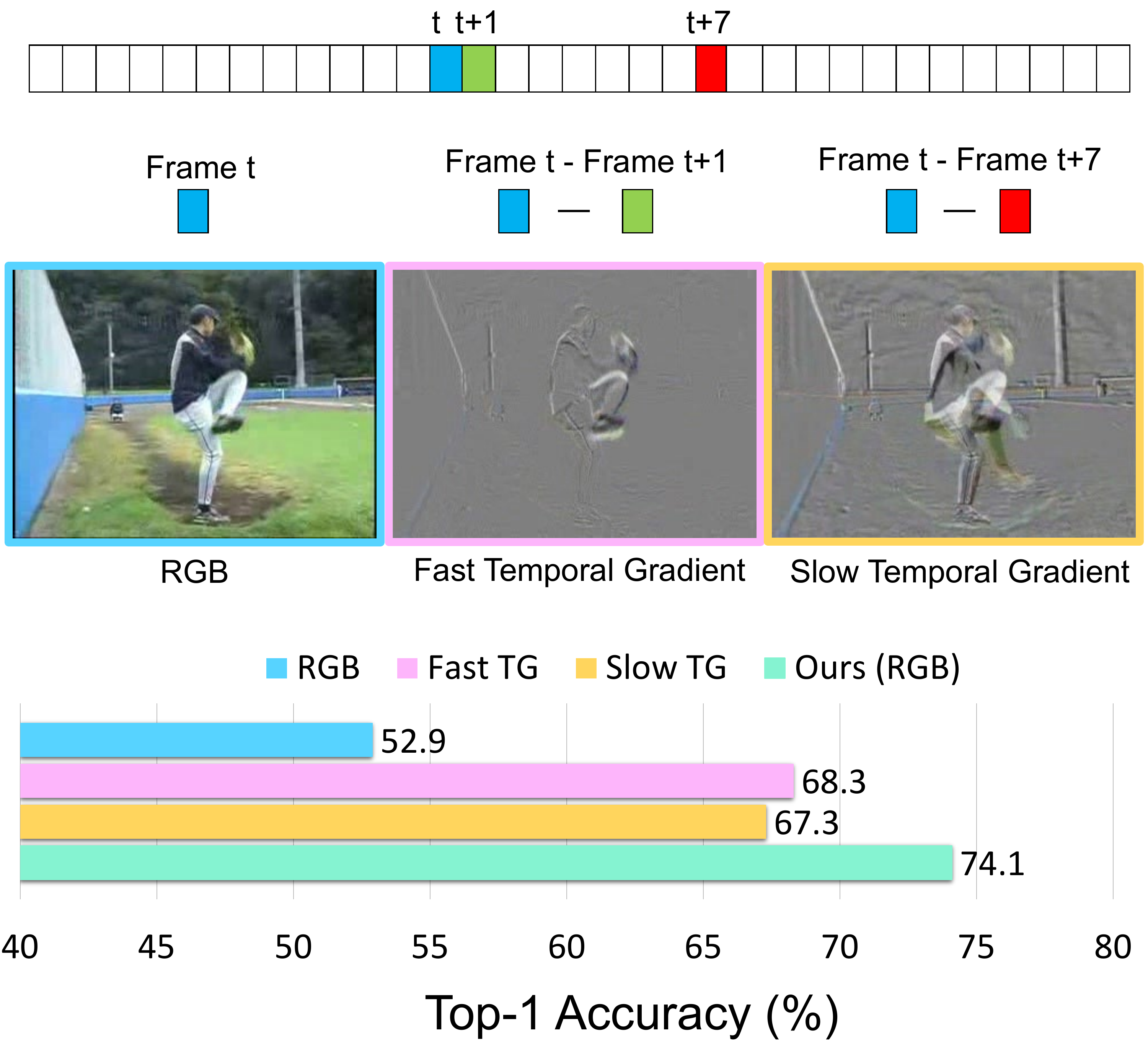}
	\caption{
	\textbf{Top:} A sketch diagram that describes the formulation of different modalities (\ie, RGB, fast and slow temporal gradient (TG)). 
	\textbf{Bottom:}
	A comparison of Top-1 accuracy with FixMatch~\cite{sohn2020fixmatch} as the baseline of semi-supervised learning method. The chart compares the performance of different input modalities (\ie, RGB, slow TG and fast TG). The remarkable performance with TG motivates us to figure out a way to efficiently utilize this fruitful modality.  By distilling knowledge from temporal gradient to RGB, our model is able to significantly outperform the models taking either temporal gradient or RGB frames as input.
	} 
	\label{fig:motivation}
	\vspace{-3mm}
\end{figure}

\section{Introduction}
\label{sec:intro}

As a fundamental task for video understanding, video action recognition has drawn much attention from the community and industry~\cite{tran2015learning, carreira2017quo, tran2018closer, feichtenhofer2019slowfast}. Unlike image-related tasks, networks for video-related tasks are normally easier to overfit due to the complexity of the tasks~\cite{tran2017convnet, tran2015learning, kataoka2020would}. The common practice is to firstly pre-train the network on large-scale datasets (\eg, Kinetics~\cite{carreira2019short} of up to $650,000$ video clips) and then finetune on downstream small datasets to obtain better performance~\cite{feichtenhofer2019slowfast,qian2021spatiotemporal,pan2021videomoco,han2020self}.

However, since annotating large-scale video datasets is time-consuming and expensive, training models on a large dataset collected with complete annotations is impeded. To utilize large-scale datasets with acceptable costs, some researchers have turned to designing semi-supervised learning models which have good generalization ability with limited annotations~\cite{jing2021videossl,singh2021semi,zou2021learning,xiong2021multiview}. As pseudo-label based methods (\eg, FixMatch~\cite{sohn2020fixmatch} and MixMatch~\cite{ berthelot2019mixmatch}) have shown outstanding performance on semi-supervised image classification, most previous video-based methods are heavily built on them to utilize the unlabeled data. Although these preliminary attempts have obtained acceptable results, most methods ~\cite{jing2021videossl,zou2021learning} are just taking video clips as `images' in 3D without further consideration of the video properties.

Videos are significantly different from images, and the key differences are the temporal information span in multiple frames and the inherent multimodal property. The temporal information refers to the motion signal between frames, and usually the features of contiguous frames from the same video change smoothly. \zzrevise{The multimodal consistency refers to the features of multiple modalities in the same video clip should be consistent since they are encoding the same content.}{The multimodal consistency expects the features extracted from the same video clip to be consistent, as they encode identical content.} Without special designs to specifically focus on temporal information and multimodal consistency, the potential of semi-supervised action recognition is not fully unleashed.

Some previous studies\cite{wang2016temporal, zhao2018recognize} introduce temporal gradient\footnote{The difference between two RGB video frames with a short interval.} as an additional modality to better utilize the temporal information encoded in videos as it is rich in motion signals. The temporal gradient can be formulated as:
$TG = x_t^{RGB} - x_{t+n}^{RGB}$,
where $x$ represents a video, $t$ denotes the frame index and $n$ denotes the interval for calculating temporal gradient.

Inspired by these studies, we made a trial with temporal gradient under the semi-supervised settings and found that a much better performance could be generated when the input frames in RGB are replaced with temporal gradient. As shown in~\Cref{fig:motivation}, the Top-1 accuracy of temporal gradient is $\sim$25\% higher than using the RGB as input on the UCF-101 dataset with only $20$\% of labeled data for training.

Why is the temporal gradient so much better than the RGB frames when the training data are limited? We hypothesize that the key is in the detailed and fine-grained motion signals encoded in the temporal gradient. The gradient along the temporal dimension is color-invariant and explicitly encodes the representative motion information of the actions in the video. This helps models generalize much easier when the labels are extremely limited. Therefore, in this paper, we propose to train a semi-supervised action recognition RGB based model to mimic both the fine-grained and high-level features from the temporal gradient.

We start from FixMatch~\cite{sohn2020fixmatch}, a typical pseudo-label based semi-supervised model, as the baseline framework. However, without any further constraints in the feature level, pseudo-label based methods perform poorly in the case of very limited labels, as many generated pseudo-labels are inaccurate.
Therefore, we propose two constraints to help the model extract temporal information in video with multiple modalities and improve the consistency between the multimodal representations. To leverage the detailed and fine-grained motion signals from temporal gradient, we propose a knowledge distillation strategy using block-wise dense alignment. It helps the student RGB model learn from the teacher temporal gradient model efficiently and effectively. To further improve high-level representation space across different modalities, we perform contrastive learning between the features from RGB and temporal gradient sequences to enforce the high-order similarity. Given the two constraints at the feature level, our proposed model is able to achieve much better performance.

Unlike the existing methods, our model has two unique advantages. First, our model requires no additional computation or parameters for inference. In the training, we distill the knowledge from temporal gradient to the RGB-based network; in the testing, only the RGB model is required. Second, our model is simple, yet effective. We conducted experiments on multiple public action recognition benchmarks including UCF-101, HMDB-51, and Kinetics-400. Our proposed method significantly outperforms all the state-of-the-art methods by a large margin.

\section{Related Work}
\noindent\textbf{Semi-supervised learning in images.} The semi-supervised image classification task has been well studied and many methods have been proposed including Pseudo-Label~\cite{lee2013pseudo},  S4L~\cite{zhai2019s4l}, MeanTeacher~\cite{tarvainen2017mean}, MixMatch~\cite{berthelot2019mixmatch}, UDA~\cite{xie2020unsupervised},  FixMatch~\cite{sohn2020fixmatch}, UPS~\cite{rizve2020defense}, etc. The Pseudo-Label~\cite{lee2013pseudo} is an early method which uses the confidence (softmax probabilities) of the unlabeled data as labels and to train the network jointly with a small ratio labeled data and much more unlabeled data. Many improved versions of Pseudo-Label have been proposed while the key is to improve the quality of the labels~\cite{sohn2020fixmatch, rizve2020defense}. Following a  state-of-the-art method on image classification---FixMatch~\cite{sohn2020fixmatch}, many FixMatch-alike methods  achieve the state-of-the-art performance on many other tasks including detection~\cite{wang20213dioumatch}, segmentation~\cite{zou2020pseudoseg}, etc. Although these methods achieve remarkable performance on image-based tasks, some recent studies show that the performances are not satisfying when directly applying these methods to video semi-supervised tasks~\cite{jing2021videossl, singh2021semi}.

\smallskip\noindent\textbf{Semi-supervised learning in videos.} Although there have been a few semi-supervised video action recognition methods~\cite{jing2021videossl,singh2021semi,zou2021learning, xiong2021multiview} proposed, most of them directly apply the image-based methods to videos with less focus on the temporal dynamics of videos. VideoSSL~\cite{jing2021videossl} made the first attempt to build a benchmark for the video semi-supervised learning task by training the network with ImageNet pre-trained models, which explicitly guides the model to learn the appearance information in each video. It also shows that the existing image-based methods (\eg Pseudo-Label~\cite{lee2013pseudo}, Mean-Teacher~\cite{tarvainen2017mean}) have inferior performance on video semi-supervised benchmarks. TCL~\cite{singh2021semi} is a recently proposed method that jointly optimizes the network by employing a self-supervised auxiliary task and a group contrastive learning. By using multimodal data, MvPL~\cite{xiong2021multiview} achieved the state-of-the-art performance by sharing the same model with different input modalities (RGB, temporal gradient, and optical flow) and generating pseudo labels with the ``confidence'' of multiple modalities. Compared with these methods, our method specifically focuses on learning the temporal information from Temporal Gradient with our proposed constraints and significantly outperforms the state-of-the-art methods on multiple public benchmarks.

\smallskip\noindent\textbf{Multimodal Video Feature Learning.} Videos could be viewed from different modalities while each modality encodes information from a unique perspective. For example, a video in general RGB couples both spatial and temporal information, the temporal gradient is color invariant which mainly encodes the difference between frames, and the optical flow explicitly encodes the motion information for each pixel. The features from different modalities are normally complementary to each other, and therefore the feature fusions are normally performed for better performance. The pioneer work is the Two-Stream~\cite{simonyan2014two, feichtenhofer2016convolutional} model which fuses features from both RGB video clips and optical flow clips. With the complementary information from different modalities, the multimodal network is able to achieve better performance~\cite{simonyan2014two, feichtenhofer2016convolutional, wang2015towards,wang2020makes,alwassel2019self}. However, there are additional computation and latency during inference. Unlike the normal multimodal feature fusion model, our model distills the motion-related representation from the temporal gradient to the base RGB model, while only the base model and RGB frames are needed during the inference stage. Moreover, our model outperforms the teacher model with only RGB as input during the inference.

\smallskip\noindent\textbf{Contrastive learning.} Contrastive learning methods have achieved remarkable performance on downstream image classification~\cite{tian2020contrastive, chen2020simple,misra2020self, he2020momentum,grill2020bootstrap,caron2020unsupervised}. The key idea is that the representation can be learned by minimizing the distance of features of positive pairs (two views of the same data sample) and maximizing the distance of features of negative pairs (two different data samples). Recently, many researchers proposed to use temporal contrastive learning for video self-supervised learning~\cite{han2020self, feichtenhofer2021large,qian2021spatiotemporal,pan2021videomoco,huang2021self}. In this paper, to better utilize the unlabeled data for semi-supervised action recognition, we propose to use the cross-modal contrastive loss to enforce the consistency of features from RGB clips and temporal gradient clips. We demonstrate that the cross-modal contrastive method is very effective for the proposed semi-supervised learning.

\section{Method}
\begin{figure*}[t]
\centering
\includegraphics[width=\textwidth]{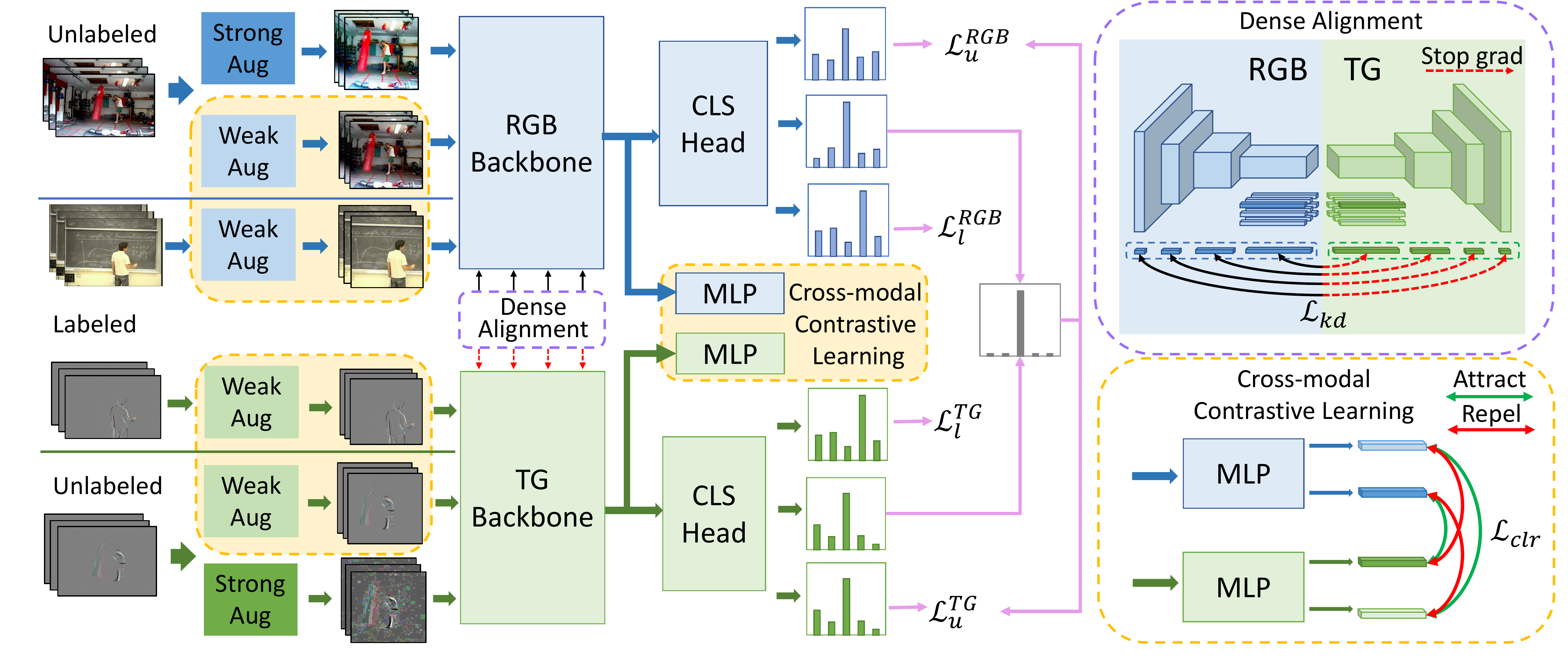}
\caption{\textbf{An overview of our proposed framework.} Our method consists of two parallel models with different input modalities (\ie, RGB and TG) of video clips. The entire framework is jointly optimized with (1) two parallel FixMatch frameworks with pseudo-labeling, (2) cross-modal dense feature alignment, and (3) cross-modal contrastive learning. }
\label{fig:overview}
\vspace{-2mm}
\end{figure*}

The objective of our method is to improve the performance on the semi-supervised action recognition task by introducing and utilizing an effective view of videos: Temporal Gradient. The overview of our proposed framework is shown in \Cref{fig:overview}, which consists of three main components: (1) the FixMatch framework with weak-strong augmentation strategy to generate better pseudo-labels for unlabeled data (2) cross-modal dense feature alignment between the features from RGB clips and TG clips for network to learn the fine-grained motion signals, and (3) cross-modal contrastive learning to learn high-level consistency feature across RGB and TG clips. The formulation for each component is introduced in the following subsections.

\subsection{FixMatch}

Considering a multi-class classification problem, we denote $\mathcal{X} = \{(x_i, y_i)\}_{i=1}^{N_l}$ as the \emph{labeled} training set, where $x_i \in \mathcal{R}^{T\times H\times W\times 3}$ is the $i$-th sampled video clip, $y_i$ is the corresponding one-hot ground truth label, and $N_l$ is the number of data points in the labeled set.
Similarly, we denote $\mathcal{U} = \{x_j\}_{j=1}^{N_u}$ as the \emph{unlabeled} set, where $N_u$ is the number of data points in the unlabeled set. We use $f_{\theta}$ to denote a classification model with trainable parameters $\theta$. We use $\alpha(\cdot)$ to represent the weak (standard) augmentation (\ie, random horizontal flip, random scaling, and random crop in video action recognition), and $\mathcal{A}(\cdot)$ to represent the strong data augmentation strategies (i.e., Randaugment~\cite{cubuk2020randaugment}).

The network $f_{\theta}$ is optimized with each video clip consisting of $T$ frames as $x_i$. 
For a mini-batch of \emph{labeled} data $\{(x_i, y_i)\}_{i=1}^{B_l}$, the network is optimized by minimizing the cross-entropy loss $\mathcal{L}_l$ as
\begin{equation}
\small
    \mathcal{L}_l = -\frac{1}{B_l} \sum_{i=1}^{B_l} 
    y_i \log f_{\theta}(\alpha(x_i)),
    \label{eq:loss_l}
\end{equation} 
where $B_l$ is the number of labeled samples in a batch.

For a mini-batch of \emph{unlabeled} data $\{x_j\}_{j=1}^{B_u}$, FixMatch enforces the model to produce consistent predictions of the same unlabeled data sample with different extent of augmentations. Specifically, pseudo labels $\hat{y}$ for the unlabeled data are usually generated via confidence thresholding as: 
\begin{equation}
\label{eq:pseudo_label}
\small
    \mathcal{C} = \{x_j | \text{max} f_{\theta}(\alpha(x_j)) \geq \gamma\},
\end{equation}
where $\gamma$ denotes a pre-defined threshold and $\mathcal{C}$ is the confident example set from a mini-batch. The confident predictions $f_{\theta}(\alpha(x_j))$ in the set $\mathcal{C}$ are then transformed into one-hot labels $\hat{y}_j$ by taking the \textit{argmax} operation. Then a cross-entropy loss $\mathcal{L}_u$ will be optimized over the samples in $\mathcal{C}$ and its generated one-hot labels as: 
\begin{equation}
\small
    \mathcal{L}_u = -\frac{1}{B_u} \sum_{x_j \in \mathcal{C}}
    \hat{y}_j \log f_{\theta}(\mathcal{A}(x_j)),
    \label{eq:loss_u}
\end{equation}
where $B_u$ is the number of unlabeled samples in a batch.

With the loss over both labeled and unlabeled data, the entire FixMatch is optimized with the objective function as: 
\begin{equation}
\small
    \mathcal{L}_{fm} = \mathcal{L}_l + \mathcal{L}_u.
\end{equation}

\subsection{Parallel Framework for Temporal Gradient}
\label{sec:parallel}

Temporal gradient (TG) $(\frac{\partial V}{\partial t})$ between two RGB frames in a video encodes the appearance change and corresponds to the temporal information that changes dynamically. Therefore, the response is accentuated by the moving objects, especially the boundaries. FixMatch\cite{sohn2020fixmatch} is originally designed for the image classification task and pays little attention to the temporal information of videos, therefore, we extend it to jointly train with RGB and TG to explicitly focus more on capturing the temporal information. To avoid additional computation and delays for processing temporal gradients during model inference on unseen videos, we propose to distill fine-grained motion signals from TG to RGB without introducing extra input or parameters for inference.

The RGB and temporal gradient information are complementary with each other. The RGB encodes spatial and temporal information in a general way, while the temporal gradient has a focus on the motion signals, as illustrated in \Cref{fig:motivation}. Therefore, for each video clip, the predictions from both RGB network and TG network are averaged and then used to generate the pseudo labels. In this way, the fused pseudo-label generation is reformulated as:
\begin{equation}
\label{eq:pseudo_label_avg}
\small
    \mathcal{C} = \{x_j | \text{max} (\frac{f_{\theta_{R}}(\alpha(x_j^{RGB})) + f_{\theta_{T}}(\alpha(x_j^{TG}))}{2} )  \geq \gamma\}.
\end{equation}

Having access to both features from RGB and TG, the quality of the fused pseudo labels are more accurate than the predictions from each model alone, and a more detailed ablation study is provided in \Cref{sec:Abla}. The fused pseudo labels will be jointly used with unlabeled data to train both the TG and RGB model. For the temporal gradient model, the training objective is also a summation of \Cref{eq:loss_l} and \Cref{eq:loss_u} but for TG.
\begin{equation}
\small
    \mathcal{L}_{fm}^{TG} = \mathcal{L}_{l}^{TG} + \lambda_u \mathcal{L}_u^{TG}.
\end{equation}

\subsection{Cross-modal Dense Feature Alignment}
\label{sec:alignemnt}

To learn detailed fine-grained motions from temporal gradient, we propose to distill the knowledge from temporal gradient model to the RGB model. The similarities between the features from both temporal gradient and RGB clips are minimized by the cross-model dense feature alignment module as: 
\begin{equation}
\label{eq:alignment obj}
\small
\min \left[\mathcal{D}\left(\mathcal{F}^{RGB}_i, \mathcal{F}^{TG}_i\right)\right],
\end{equation} where $\mathcal{F}^{RGB}_i , \mathcal{F}^{TG}_i\in \mathbb{R}^{C_i \times T_i \times H_i \times W_i}$ denote the output features of the $i$-th block in the RGB and TG models, and $\mathcal{D}$ represents a pairwise function evaluating the representation differences. There are many choices for $\mathcal{D}$ and we experiment with three different functions: L1, L2 and Cosine Similarity losses (shown in \Cref{eq:alignment func}, where $\|\cdot\|_{1}$ and $\|\cdot\|_{2}$ are $\ell_{1}/\ell_{2}$-norm). A more detailed discussion is provided in \Cref{sec:Abla}.
\begin{align}
\label{eq:alignment func}
\small
\begin{split}
\mathcal{D}_{L1}\left(\mathcal{F}_{1}, \mathcal{F}_{2}\right)&={\left\|\mathcal{F}_{1}-\mathcal{F}_{2}\right\|_{1}},
\\
\mathcal{D}_{L2}\left(\mathcal{F}_{1}, \mathcal{F}_{2}\right)&={\left\|\mathcal{F}_{1}-\mathcal{F}_{2}\right\|_{2}},
\\
\mathcal{D}_{cos}\left(\mathcal{F}_{1}, \mathcal{F}_{2}\right)&=-\frac{\mathcal{F}_{1}}{\left\|\mathcal{F}_{1}\right\|_{2}} \cdot \frac{\mathcal{F}_{2}}{\left\|\mathcal{F}_{2}\right\|_{2}}.
\end{split}
\end{align}

An key setting in our online knowledge distillation method is the stop-gradient $(stopgrad)$ operation on the temporal gradient side, which means the teacher model would not receive any gradient from the alignment loss. This helps the TG model avoid degeneration by the alignment with the RGB student model. As shown in \Cref{eq:alignment loss}, the alignment loss term for learning fine-grained motion features is:
\begin{equation}
\label{eq:alignment loss}
\small
\mathcal{L}_{kd} = \left[\mathcal{D}\left(\mathcal{F}^{RGB}_i, stopgrad(\mathcal{F}^{TG}_i)\right)\right].
\end{equation}

\subsection{Cross-modal Contrastive Learning}
\label{sec:contrastive}

The dense feature alignment explicitly enables the RGB network to mimic the fine-grained motion signals from temporal gradient. We hypothesize that the global high-level representations across different modalities are also valuable and crucial. Therefore, cross-modal contrastive learning is employed as another module to discover the mutual information that coexists in both TG and RGB clips.
 Following the principle of SimCLR\cite{chen2020simple} and CMC\cite{tian2020contrastive}, we form the contrastive learning with positive pairs and negative pairs. Specifically, we consider the two modalities of the same video clip as a positive pair $\{k^{+}\}$ and the two modalities of different video clips as negative pairs $\{k^{-}\}$. The learning objective is to maximize the similarity of positive pairs and minimize the similarity of negative ones. We adopt \mbox{InfoNCE} loss \cite{oord2018representation} as the objective function over the features extracted from RGB and TG:
\begin{equation}\label{eq:infonce} 	\small
	\mathcal{L}_{clr} = -\log{\frac{ {\sum_{k \in \{k^+\}}} \exp\left({\mathrm{sim}(q, k) / \tau}\right)}{ {\sum_{k \in \{k^+, k^-\}}} {\exp\left({\mathrm{sim}(q, k) / \tau}\right)}   }},
\end{equation} with $\tau$ being a temperature hyper-parameter for scaling.
All embeddings are $\ell_2$  normalized and dot product (cosine) similarity is used to compare them $\mathrm{sim}(q, k) = q^\top k / \lVert q\rVert \lVert k\rVert$.
	
It is worth noting that this cross-modal contrastive learning directly uses all weakly augmented samples of the two modalities ($\alpha(x^{RGB/TG}_i)$) in the FixMatch, including both labeled (the labels are not used) and unlabeled data. Therefore, there is no additional computation for the data loading and preprocessing.

\smallskip\noindent\textbf{Total Loss}:
Our entire model based is jointly trained with cross-entropy loss over labeled data,cross-entropy loss over the unlabeled data with pseudo-labels, the dense alignment over both labeled and unlabeled data, and the cross-modal contrastive loss over both labeled and unlabeled data. Overall, the final objective function of our method is:
\begin{equation}
\small
    \mathcal{L}_{total} =w_{fm} (\mathcal{L}_{fm}^{RGB} + \mathcal{L}_{fm}^{TG}) + w_{kd}\mathcal{L}_{kd} + w_{clr}\mathcal{L}_{clr}.
\end{equation}


\begin{table*}[htbp]
		\centering
		\tablestyle{6pt}{1.1}
		\small
        \begin{tabular}{cc|cc|cc|cc|cc|cc}
        \shline
         && \multicolumn{4}{c|}{Kinetics-400} & \multicolumn{4}{c|}{UCF-101} & \multicolumn{2}{c}{HMDB-51}
        \\ \hline
        &            & \multicolumn{2}{c|}{1\%} & \multicolumn{2}{c|}{10\%} & \multicolumn{2}{c|}{10\%}                     & \multicolumn{2}{c|}{20\%}
                             & \multicolumn{2}{c}{50\%}
                             \\ \shline
        Alignment & Contrast                           & Top-1         & Top-5          & Top-1          & Top-5          & Top-1                     & Top-5                     & Top-1                     & Top-5
        & Top-1                     & Top-5   
        \\ \shline
        \xmark       &     \xmark              & 5.4          & 17.0          & 40.2          & 65.4          & {38.4} & {64.8} & {54.1} & {78.1} 
        & 37.8 &	68.6
        \\
        \cmark & \xmark & 9.4          & 25.5          & 43.5          & 68.8          & 60.4                     & 84.4                     & 74.6                     & 91.7  &47.3 &74.8                   \\
        \xmark & \cmark & 5.2 & 23.1 & 42.6 & 67.4 & 58.0                     & 82.5                     & 68.6         & 89.2
        & 46.1 & 73.8
        \\
        \cmark & \cmark &\textbf{9.8} & \textbf{26.0} & \textbf{43.8} & \textbf{69.2} & \textbf{62.4}            & \textbf{84.9}            & \textbf{76.1}            & \textbf{92.1}
        &  \textbf{48.4}	& \textbf{75.9}
        \\   
        \shline 
        \end{tabular}
		\centering
		\caption{\textbf{Effectiveness of the cross-modal alignment and contrastive learning.} The results are evaluated on the validation sets. The first row shows the results of the FixMatch baseline model without any proposed modules.}
		\label{tab:abla_major}
		\vspace{-4mm}
	\end{table*}

\section{Experimental Results}
\subsection{Datasets and Evaluation}
\noindent\textbf{Datasets.} Following previous state-of-the-art semi-supervised video action recognition methods~\cite{jing2021videossl,zou2021learning, xiong2021multiview}, we evaluate our method on three public action recognition benchmarks: UCF-101~\cite{soomro2012ucf101}, HMDB-51~\cite{kuehne2011hmdb}, and Kinetics-400~\cite{kay2017kinetics}. UCF-101 is a widely used dataset which consists of $13,320$ videos belonging to $101$ classes. HMDB-51 is a smaller dataset which consists of $6,766$ videos with $51$ classes. For UCF-101 and HMDB-51, we follow the data splits that released by VideoSSL\cite{jing2021videossl}. The Kinetics-400 dataset is a large-scale dataset
consisting of $\sim$235k training videos and $\sim$20k validation videos belonging to 400 classes. For Kinetics-400, we follow the most recent state-of-the-art method MvPL~\cite{xiong2021multiview} by forming two balanced labeled subsets by randomly sampling 6 and 60 videos per class for 1\% and 10\% settings.

\smallskip\noindent\textbf{Evaluation.}
We report Top-1 accuracy for major comparisons and Top-5 accuracy for some ablation studies.

\subsection{Implementation Details}
\label{sec:implementation}
\noindent\textbf{Network architecture.} For a fair comparison with the state-of-the-art methods~\cite{jing2021videossl, xiong2021multiview}, the FixMatch~\cite{sohn2020fixmatch} framework is used as the backbone model while the 3D ResNet-18~\cite{tran2018closer, he2016deep} is adopted as feature extractors for both RGB and TG (\Cref{sec:parallel}) modalities. For each feature extractor, two individual contrastive heads with 3-layer non-linear MLP architecture are added for the cross-modal contrastive learning (\Cref{sec:contrastive}).

\smallskip\noindent\textbf{Video augmentations.} There are two types of data augmentations: weak augmentation and strong augmentation. For the weak augmentation, the random horizontal flipping, random scaling, and random cropping following \cite{zou2021learning}. To be specific, given a video clip, we firstly resize the video making the short side be $256$, and then a randomly resized crop operation is performed. The cropped clips are then resized to 224$\times$224 pixels and flipped horizontally with a $50$\% probability. For strong augmentation, the RandAugment~\cite{cubuk2020randaugment} is chosen which randomly selects a small set of transformations from a large augmentation pools (\eg, rotation, color inversion, translation, contrast adjustment, etc.) for each sample and then performs the selected data augmentation over the samples. It is worth noting that both the teacher (TG) and student (RGB) share the same weak augmentation (\ie, the inputs are identically cropped in the same area and both flipped or not). This provides a direct positional information matching, which plays a crucial role for the dense alignment in \Cref{sec:parallel}.

\smallskip\noindent\textbf{Training details.} All experiments are done with an initial learning rate of 0.2 on 8 GPUs by following the settings in~\cite{feichtenhofer2019slowfast,zou2021learning, xiong2021multiview} using the cosine learning rate decaying scheduler~\cite{loshchilov2016sgdr} and also a linear warm-up strategy \cite{goyal2017accurate}. We use momentum of 0.9 and weight decay of 10$^\text{-4}$. Dropout \cite{srivastava2014dropout} of 0.5 is used before the final classifier layer to reduce the over-fitting. Following~\cite{zou2021learning}, each mini-batch consists of 5 labeled data clips and 5 unlabeled data clips, while each input clip consists of 8 frames with a sampling stride of 8, which covers 64 frames of the raw video. We consistently train our models with 180 and 360 epochs for all experiments on UCF-101 and HMDB-51, while 45 (1\%) and 90 (10\%) epochs are trained for Kinetics-400. More training details are provided in the supplementary material. For the pseudo-label threshold, we follow~\cite{xiong2021multiview} which sets it to 0.3 for getting more training samples.
For the loss weights, $w_{fm}$ is set to 0.5 while $w_{kd}$ and $w_{clr}$ are set to 1.

\smallskip\noindent\textbf{Inference.} 
Following the recent state-of-the-art methods \cite{feichtenhofer2019slowfast, zou2021learning, xiong2021multiview}, 10 clips are uniformly sampled for each video along its temporal axis and each clip is taken 3 crops of 256$\times$256. A total of 3x10 crops are evaluated for each video.

\subsection{Effectiveness of the Cross-modal Dense Alignment and Contrastive Learning}

We begin with a direct comparison to examine our hypothesis:
\emph{multimodal constraints on local and global features can serve as two complementary extensions to existing semi-supervised methods} (FixMatch~\cite{sohn2020fixmatch} as the baseline).
To this end, our dense alignment (\Cref{sec:alignemnt}) is devised to regularize the local features, and our contrastive loss (\Cref{sec:contrastive}) is developed to distinguish global features. 
For a fair comparison, we have ablated four experimental settings (detailed in \Cref{tab:abla_major}): (1) none, (2) alignment-only, (3) contrast-only, and (4) both.
Kinetics-400, UCF-101, and HMDB-51 with different labeled data ratios (\ie, 1\%, 10\%, 20\%, and 50\%) are used to ensure the generalizability of the following observations. \textit{First}, FixMatch (none) exhibits acceptable but worse performance than its three counterparts, suggesting that pseudo labeling only is inadequate when using very limited labeled data. 
\textit{Second}, dense alignment significantly elevates the performance (more than contrast-only), indicating that the fine-grained motion signal across multimodal plays an essential role in semi-supervised action recognition. \textit{Third}, introducing contrastive loss across RGB and TG modalities improves Top-1/Top-5 accuracy, revealing that global consistency in different modalities is advantageous.
\textit{Finally}, dense alignment and contrastive loss enforce the model learning from complementary perspectives because implementing both on top of FixMatch surpasses either one of them.
We hope that our discovery on multimodal constraints can shed new light on semi-supervised action recognition in video analysis.

\zzrevise{}{
\smallskip\noindent\textbf{Overfitting is alleviated.}  \Cref{tab:overfitting} (Suppl.) presents a significant accuracy gap between the training and testing set, showing that FixMatch severely overfits to the training set. Our method effectively reduces the gap by imposing additional regularization on models with RGB as input.
}

\begin{table*}[!t]
\centering

\begin{threeparttable}
\resizebox{\textwidth}{!}{
\begin{tabular}{lccccccccccccc}
\toprule
 & w/ ImageNet           &          & \multicolumn{2}{c}{Kinetics-400} &           & \multicolumn{4}{c}{UCF-101}                                   &  & \multicolumn{3}{c}{HMDB-51}                   \\
\cmidrule{4-5}
\cmidrule{7-10}
\cmidrule{12-14}
Method                                           & distillation & Backbone    & 1\%              & 10\%              &           & 5\%             & 10\%            & 20\%            & 50\%            &  & 40\%            & 50\%            & 60\%            \\ \midrule
Pseudo-Label~\cite{lee2013pseudo} (ICMLW 2013)   & \xmark       & R3D-18      & 6.3            & -               &           & 17.6          & 24.7          & 37.0          & 47.5          &  & 27.3          & 32.4          & 33.5          \\
MeanTeacher~\cite{tarvainen2017mean} (NIPS 2017) & \xmark       & R3D-18      & 6.8            & 19.5            &           & 17.5          & 25.6          & 36.3          & 45.8          &  & 27.2          & 30.4          & 32.2          \\
S4L~\cite{zhai2019s4l} (ICCV 2019)               & \xmark       & R3D-18      & 6.3            & -               &           & 22.7          & 29.1          & 37.7          & 47.9          &  & 29.8          & 31.0          & 35.6          \\
UPS~\cite{rizve2020defense} (ICLR 2021)          & \xmark       & R3D-18      & -              & -               &           & -             & -             & 39.4          & 50.2          &  & -             & -             & -             \\ \midrule
VideoSSL~\cite{jing2021videossl} (WACV 2021)     & \cmark       & R3D-18      & -              & 33.8            &           & 32.4          & 42.0          & 48.7          & 54.3          &  & 32.7          & 36.2          & 37.0          \\
TCL~\cite{singh2021semi} (CVPR 2021)             & \xmark       & R3D(TSM)-18 & 7.7            & -               &           & -             & -             & -             & -             &  & -             & -             & -             \\
ActorCutMix~\cite{zou2021learning} (arXiv 2021)  & \xmark       & R(2+1)D-34  & -              & -               &           & 27.0          & 40.2          & 51.7          & 59.9          &  & 32.9          & 38.2          & 38.9          \\
MvPL*~\cite{xiong2021multiview} (arXiv 2021)     & \xmark       & R3D-18      & 5.0            & 36.9            &           & 41.2          & 55.5          & 64.7          & 65.6          &  & 30.5          & 33.9          & 35.8          \\
\textbf{Ours}                                    & \xmark       & R3D-18      & \textbf{9.8}   & \textbf{43.8}   & \textbf{} & \textbf{44.8} & \textbf{62.4} & \textbf{76.1} & \textbf{79.3} &  & \textbf{46.5} & \textbf{48.4} & \textbf{49.7}

\\
\bottomrule
\end{tabular}
}
\begin{tablenotes}
      \small
      \item * indicates the method is reimplemented by ourselves. The input modalities are RGB and TG.
    \end{tablenotes}
\end{threeparttable}
\caption{\textbf{Comparison with the state-of-the-arts methods.}
The results are reported with Top-1 accuracy (\%) on the validation sets.
The best performance of each setting is in \textbf{bold}.
}
\vspace{-5mm}
\label{tab:main_all}
\end{table*}
\subsection{Comparison with State-of-the-art Methods}

To demonstrate the capability and potential of our proposed method, we compared with the most recent state-of-the-art methods for the semi-supervised action recognition task on public datasets including Kinetics-400, UCF-101 and HMDB-51. As shown in \Cref{tab:main_all}, we mainly compare with two types of methods including image-based methods~\cite{lee2013pseudo, tarvainen2017mean, zhai2019s4l} which were originally designed for image classification and then simply adopted to video tasks and video based methods ~\cite{jing2021videossl, singh2021semi, zou2021learning, xiong2021multiview} which were specifically designed for video action recognition task.

\smallskip\noindent\textbf{Comparison with image-based methods.}
The first three rows in \Cref{tab:main_all} show the results of image-based methods including \ie, Pseudo-Label\cite{lee2013pseudo}, MeanTeacher\cite{tarvainen2017mean} and S4L\cite{zhai2019s4l}. In general, the results of all the three image-based methods across over the three datasets with all different labeled percentages are much lower than the results of all video-based based methods. This confirms that it is necessary to propose methods specifically designed based on the video temporal and multimodal attributes.

\smallskip\noindent\textbf{Comparison with video-based methods.} The overall performance of the video-based performance are much higher. VideoSSL surpasses all the image-based methods by using Imagenet pre-trained model to guide the learning, and TCL~\cite{singh2021semi} use self-supervised learning task as auxiliary task and the group contrastive for the video semi-supervised learning. Both the ActorCutMix~\cite{zou2021learning} and MvPL~\cite{xiong2021multiview} are adapted from FixMatch~\cite{sohn2020fixmatch}. Benefited by our proposed cross-modal dense alignment and cross-modal contrastive, our method outperforms all these methods by a significant margin on three datasets under all the experimental settings (different ratio of labels).

\subsection{Ablation Studies}
\label{sec:Abla}

To understand the impact of each part of the design in our method, we conduct extensive ablation studies on the UCF-101 dataset with 20\% labeled setting.

\begin{figure}[t]
	\centering
    \hspace*{-0.5cm} 
	\includegraphics[width=1\columnwidth]{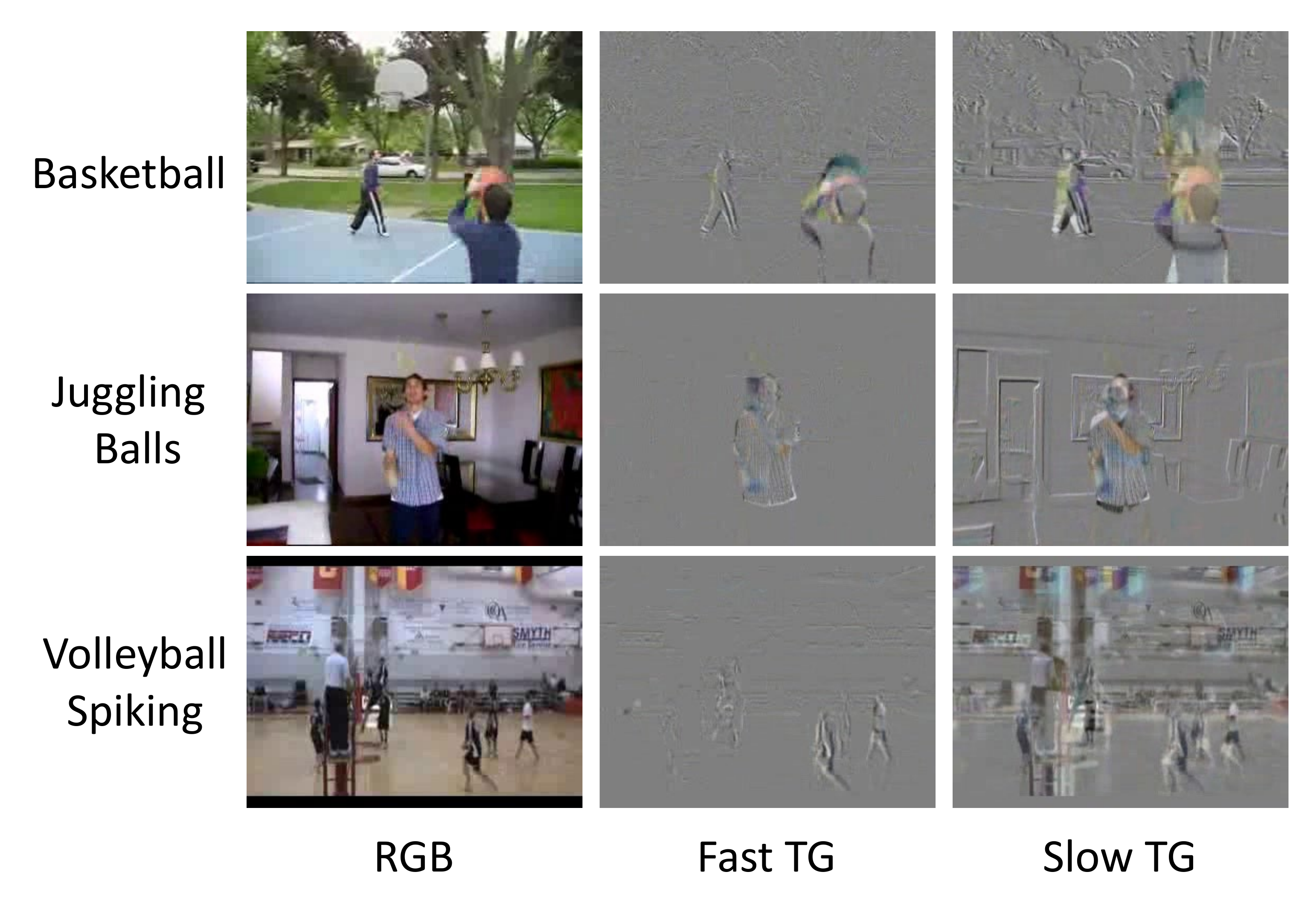}
	\caption{\textbf{Visualization of the slow and fast temporal gradient.} Slow temporal gradient contains a more noisy background of the shooting environment while fast temporal gradient focuses more on the activity-related moving objects.} 
	\label{fig:compare_slowfast_tg}
	\vspace{-5mm}
\end{figure}

\smallskip\noindent\textbf{Fast temporal gradient is better.}
Temporal gradient (TG) is calculated by differing two RGB frames and the stride of them could be small or large to generate either fast or slow TG. To delve deeper into the effect of different strides, we conduct experiments with fast TG (calculation stride = 1) and slow TG (calculation stride = 7), and the results are shown in \Cref{tab:abla_fast}. The first group compare the performance with the baseline FixMatch framework with different modalities of data as input. The results confirm that both the slow TG and fast TG perform much better than RGB (more than 25\% higher), and also demonstrate that the Fast TG is better than the slower TG for the semi-supervised setting. The second group of \Cref{tab:abla_fast} compares the final performance of our model with different temporal gradients. When the pseudo-labels are generated by the fast TG, the model beats the performance with slow TG with a large margin (74.1\% vs. 68.2\%). To figure out the reason why the performance of fast TG is much higher than slower TG, we visualized the two types of temporal gradient for three video clips and the visualization are shown in \Cref{fig:compare_slowfast_tg}. The comparison shows that the slow TG has much noisy background information especially when the cameras have significant movement while the fast temporal gradient information focuses more on the boundary of the fast moving objects (\eg, people, balls). Both the quantitative and qualitative results 
verify the advantages of the fast TG over the slower TG for semi-supervised action recognition. 

\begin{table*}[t]
    \captionsetup[subtable]{font=normalsize}

	\begin{subtable}[t]{0.25\textwidth}
		\centering
		\tablestyle{2pt}{1.02}
		\small
		\begin{tabular}{cc|c}
Student & Teacher & Top-1          \\ \shline
RGB     & -       & 52.9          \\
Slow TG & -       & 67.3          \\
Fast TG & -       & \textbf{68.3} \\ \hline
RGB     & Slow TG  & 68.2          \\
RGB     & Fast TG  & \textbf{74.1}
\end{tabular}
		\caption{\textbf{Fast temporal gradient is better.} 
	}
		\label{tab:abla_fast}
	\end{subtable}
	\begin{subtable}[t]{0.25\textwidth}
		\tablestyle{2pt}{1.02}
		\small
		\begin{tabular}{c|cc}
     Align. Loss  & Top-1 & Top-5 \\ \shline
     
- & 54.1 & 78.1 \\ \hline
L1     & 74.0 & 91.3 \\
L2     & 74.4 & 91.4 \\
Cosine & \textbf{74.6} & \textbf{91.7}
\end{tabular}
\vspace{6pt}
		\caption{\textbf{Dense alignment functions.}}
		\label{tab:abla_alignloss}
	\end{subtable}
	\begin{subtable}[t]{0.2\textwidth}
		\tablestyle{2pt}{1.02}
		\small
		\begin{tabular}{c|cc}
        Stopgrad & Top-1          & Top-5          \\ \shline
        \xmark        & 60.0          & 84.4          \\
        \cmark        & \textbf{74.6} & \textbf{91.7}
        \end{tabular}
        \vspace{17pt}
		\caption{\textbf{Stop gradient in knowledge distillation.}}
		\label{tab:abla_stopgrad}
	\end{subtable}
	\begin{subtable}[t]{0.3\textwidth}
		\tablestyle{2pt}{1.02}
		\small
		\begin{tabular}{c|cc}
        Pseudo-label Metric & Top-1          & Top-5          \\ \shline
        RGB                 & 73.6          & 91.0          \\
        TG                  & 74.1          & 91.3          \\
        Self          & 72.8          & 91.6          \\
        Average         & \textbf{74.6} & \textbf{91.7}
        \end{tabular}
        \vspace{6pt}
		\caption{\textbf{Metrics for the pseudo-labels.}}
		\label{tab:abla_pl_metric}
	\end{subtable}

	\begin{subtable}[t]{0.35\textwidth}
		\centering
		\tablestyle{2pt}{1.02}
		\small
		\begin{tabular}{cccc|cc}
\multicolumn{4}{c|}{Aligned Block Index} & \multicolumn{2}{c}{Accuracy}  \\ \hline
1st     & 2nd     & 3rd     & 4th & Top-1          & Top-5          \\ \shline
\xmark  & \xmark  & \xmark  & \cmark  & 71.4          & 90.2          \\
\xmark  & \xmark  & \cmark  & \cmark  & 74.0          & 91.4          \\
\xmark  & \cmark  & \cmark  & \cmark  & 74.4          & \textbf{91.8} \\
\cmark  & \cmark  & \cmark  & \cmark  & \textbf{74.6} & 91.7         
\end{tabular}
		\caption{\textbf{Align them in block-wise.} }
		\label{tab:abla_blockwise}
	\end{subtable}
	\begin{subtable}[t]{0.3\textwidth}
	\centering
		\tablestyle{2pt}{1.02}
		\small
		\begin{tabular}{l|cc}
                   & Top-1          & Top-5          \\ \shline
    Plain               & 71.1          & 90.0          \\
    + LR warm-up          & 71.9          & 91.1          \\
    + Sup. warm-up & 74.1          & 91.0          \\
    + PreciseBN         & \textbf{74.6} & \textbf{91.7}
    \end{tabular}
    \vspace{5pt}
    \caption{\textbf{The crucial training tricks.} 
	}
	\label{tab:abla_tricks}
	\end{subtable}
	\begin{subtable}[t]{0.35\textwidth}
	\centering
		\tablestyle{2pt}{1.02}
		\small
		\begin{tabular}{c|cc}
        Tempature $\tau$ & Top-1          & Top-5 \\ \shline
        0.1              & 74.8          & 91.8 \\
        0.2              & 75.2          & \textbf{92.4} \\
        0.5              & \textbf{76.1} & 92.1 \\
        1.0              & 74.3          & 92.2
        \end{tabular}
        \vspace{5pt}
        \caption{\textbf{Ablation on contrastive temperature.}}
        \label{tab:abla_temperature}
	\end{subtable}

	\caption{\textbf{Ablation studies} on UCF101 split-1 under 20\% semi-supervised setting  (only use 20\% labeled data). The results are reported with Top-1 and Top-5 accuracy on the validation set. Backbone: 3D ResNet-18~\cite{he2016deep,tran2018closer}, each input clip consists of 8 frames sampled from a single video with the inter-frame interval of 8. Except for the study (a), all the other results are evaluated with PreciseBN. Except for the study (g), all the other experiments are without cross-modal contrastive learning for better comparisons.}
	\vspace{-2mm}
	\vspace{-5pt}
	\label{tab:ablations}
\end{table*}

\smallskip\noindent\textbf{The choice of alignment functions.} As discussed in \Cref{sec:alignemnt}, there are many possible choices for the alignment loss function as long as it can effectively enforce the similarity between the two features. Here we studied the performance of three different alignment functions including L1, L2 and Cosine Similarity loss. As shown in \Cref{fig:compare_slowfast_tg} (b), all the three loss functions in alignment achieve high performance while Cosine Similarity (74.6\%) outperforms the other two functions (74.0\% \& 74.4\%). A possible explanation is that L1 and L2 have more strict constraints on the scale of two representations, while the Cosine Similarity loss focuses on the vector orientation (\eg, L1 and L2 losses of $\Vec{v_1}$=(10,10,10) and $\Vec{v_2}$=(1,1,1) are large while the Cosine Similarity loss is 0). Although TG is normalized to the 0-255 range during training, there is still a gap in the scales between the representations of RGB and TG. A strict constraint like L1 or L2 would have negative effects on the model for learning motion features.

\smallskip\noindent\textbf{Stop gradient in knowledge distillation.} The stop-gradient operation on the TG side stated in \Cref{sec:alignemnt} is one of the keys to the successful knowledge distillation with dense alignment. However, as the student RGB has much appearance information which TG does not have, directly training  with the dense alignment strategy would make the teacher TG model degenerate greatly and hard to focus on extracting the fine-grained motion features. The stop-gradient avoids the fine-grained motion-related representations in TG model can be disturbed by the RGB model. As shown in \Cref{tab:abla_stopgrad}, there is a 14.6\% performance drop on Top-1 accuracy (60.0\% vs. 74.6\%) when stop gradient is taken off.

\smallskip\noindent\textbf{How to generate pseudo-labels?} There are multiple ways to generate pseudo-labels since our model takes two input modalities. We compare the performance of four settings: 1) use the prediction from RGB model as pseudo-labels, 2) use the prediction from TG model as pseudo-labels, 3) each model uses the probabilities of its self-modality, and 4) fuse the results from both RGB and TG as pseudo-labels. \Cref{tab:abla_pl_metric} shows that the fused pseudo-labels are more reliable and achieve the best performance benefiting from comprehensive information of both RGB and TB.

\smallskip\noindent\textbf{Dense alignment in block-wise.} An intuitive question about our knowledge distillation framework is that which block or blocks should be densely aligned. Therefore, we conduct this ablation study by adding dense alignment to different positions (\ie, blocks) and the results are shown in \Cref{tab:abla_blockwise}. As the common practice of previous knowledge distillation methods \cite{hinton2015distilling,thoker2019cross,stroud2020d3d} is to align the high-level features of the last layers. Therefore, we start to add the dense alignment module over the features from the last (4-th) block (ResNet basic block) and then experiment with more blocks. Their performances are consistently improved when more blocks are densely aligned and the best Top-1 accuracy is achieved with all blocks aligned. Compared with the baseline, our block-wise dense alignment strategy gains a considerable improvement of 20.5\% (54.1\% to 74.6\%) which demonstrates that fine-grained motion signals are better at semi-supervised model generalization.

\smallskip\noindent\textbf{Crucial training tricks.} Through the extensive experiments, we identified several training tricks which are essential to lead to the high performance. \Cref{tab:abla_tricks} shows the impact of learning rate warm-up \cite{goyal2017accurate}, supervised warm-up\cite{xiong2021multiview} and PreciseBN\cite{wu2021rethinking}. All the three tricks could make a decent improvement, while the supervised warm-up (training with only the labeled data at the first several epochs) is the most effective one which gains an improvement of 2.7\% (71.9\% to 74.6\%). This shows that the supervised warm-up can alleviate the cold-start issue that low-quality pseudo-labels would be generated at the beginning. The performance of the semi-supervised learning model could easily have large variations~\cite{singh2021semi,oliver2018realistic,wang20213dioumatch}. These three tricks can solidly improve the performance while in the meantime make the training more stable.

\smallskip\noindent\textbf{Contrastive temperature.}
An appropriate temperature is important to the good performance of contrastive learning~\cite{chen2020simple}, we ablate the contrastive loss temperature in \Cref{eq:infonce}. As shown in \Cref{tab:abla_temperature}, a modest temperature (e.g., 0.2 or 0.5) could help the proposed cross-modal contrastive learning work better while a large (1.0) or small (0.1) temperature is not that optimal.


\section{Conclusion}

This paper has presented a novel semi-supervised learning method which introduces temporal gradient for the fruitful motion-related information and extra representation consistency crossing multiple modalities. Our proposed method uses the block-wise dense alignment strategy and cross-modal contrastive learning. Without additional computation or delay during inference, our method substantially outperforms all prior methods while achieving state-of-the-art performance on UCF-101, HMDB-51, and Kinetics-400 datasets with all the experimented settings (different labeled ratios). In the future, we plan to study the effectiveness of temporal gradient on other video-based tasks and to automatically search or generate powerful modalities.

\smallskip\noindent\textbf{Acknowledgments:} This work was supported by the National Science Foundation under Grant No. NSF-1763705.

{\small
\bibliographystyle{ieee_fullname}
\bibliography{ref}
}
\clearpage
\appendix

This document contains the supplementary materials for ``Learning from Temporal Gradient for Semi-supervised Action Recognition". It covers the implementation details (\S\ref{sec:implementation_details}), robustness evaluation with multiple types of corruptions (\S\ref{sec:robustness}), the visualization of attention maps with Grad-CAM (\S\ref{sec:gradcam}), t-SNE feature visualization (\S\ref{sec:tsne}) and an analysis of the overfitting issue (\S\ref{sec:overfitting}).

\section{Additional Implementation Details} \label{sec:implementation_details}
	
\noindent\textbf{Network architecture.}  The details of the 3D ResNet-18~\cite{tran2018closer, he2016deep} backbone architecture are illustrated in \Cref{tab:backbone_arch}. This backbone is adopted as the feature extractor for both RGB and TG modalities. There are two heads following each backbone, one is for the general classification prediction with Softmax activation (Global Average Pooling + Dropout + FC) and the other is for the projection in the contrastive learning framework (3-layer non-linear MLP with BatchNorm~\cite{ioffe2015batch} and ReLU~\cite{hahnloser2000digital,nair2010rectified}).

    \begin{table}[htbp]
        \centering

        \resizebox{\columnwidth}{!}{%
        \begin{tabular}{c|c|c}
            \shline
            
            {Layer Name}  & 
            {Output Size}& R3D-18  \\ \hline
            conv1 & L \x 56 \x 56& $3 \times 7 \times 7, 64$, stride $1 \times 2 \times 2$ \\ \hline
            \multirow{3}{*}{conv2\_x}& \multirow{3}{*}{L \x 56 \x 56} &  \multirow{3}{*}{\residual{64}{2}} \\
            &  & \\
			&  & \\ \hline
			
			\multirow{3}{*}{conv3\_x}& \multirow{3}{*}{$\frac{L}{2}$ \x 28 \x 28} &  \multirow{3}{*}{\residual{128}{2}} \\
            &  & \\
			&  & \\ \hline
			
            \multirow{3}{*}{conv4\_x}& \multirow{3}{*}{$\frac{L}{4}$ \x 14 \x 14} &  \multirow{3}{*}{\residual{256}{2}} \\
            &  & \\
			&  & \\ \hline
            \multirow{3}{*}{conv5\_x}& \multirow{3}{*}{$\frac{L}{8}$ \x 7 \x 7} &  \multirow{3}{*}{\residual{512}{2}} \\
            &  & \\
			&  & \\ \shline
        \end{tabular}}
        \captionsetup{width=\linewidth}
        \caption{
          \textbf{Backbone architecture.} Residual blocks are in brackets.
        }
        \label{tab:backbone_arch}
    \end{table}

\smallskip\noindent\textbf{Video Augmentations.} We implement our method using MMAction2\footnote{MMAction2: \href{https://github.com/open-mmlab/mmaction2}{https://github.com/open-mmlab/mmaction2}}~\cite{2020mmaction2}. For weak augmentation, we use the \textit{Resize}, \textit{RandomResizedCrop}, and \textit{Flip} in MMAction2. For strong augmentation, we use the RandAugment~\cite{cubuk2020randaugment} implemented with imgaug~\cite{imgaug} . 

\smallskip\noindent\textbf{Temporal Gradient Normalization.}
Following Xiong~\etal~\cite{xiong2021multiview}, we normalize the temporal gradient to fit the common 0-255 range by adding 255 and dividing by 2.

\section{Robustness Against Input Corruptions} 
	\label{sec:robustness}
	
    To verify the hypothesis that our method learns more motion-related features from the temporal gradient and is more robust to contrast and brightness variations, we evaluate the models with different corruptions (\ie, random contrast adjustment noise, random brightness adjustment noise and conversion to grayscale) during the testing stage. As shown in \Cref{tab:robustness}, our method is more robust than the baseline to all types of corruptions. It is worth noting that in the gray-scale corruption case (the inputs lose all color information), the performance of baseline drops 28.0\% (51.8\% relative) while ours only drops 14.6\% (19.2\% relative). 
    
    \begin{table}[htbp] 
    \centering
    \begin{tabular}{l|cc}\shline
    Corruptions      & Baseline & Ours \\ \hline
    No Corruption    & 54.1     & 76.1 \\ \hline
    Contrast Noise   & 53.1 \red{(-1.0)}    & 75.3  \red{(-0.8)}\\
    Brightness Noise & 52.2 \red{(-1.9)}     & 75.2 \red{(-0.9)} \\
    Grayscale      & ~~26.1 \red{(-28.0)}     & ~~61.5 \red{(-14.6)}\\\shline
    \end{tabular}
    \caption{\textbf{Robustness evaluation with different corruptions.} The Contrast Noise and Brightness Noise are implemented with the \textit{EnhanceContrast} and \textit{EnhanceBrightness} of imgaug~\cite{imgaug}. All results are reported in Top-1 accuracy. The models are trained with 20\% labels (UCF101-20\%).} 
    \label{tab:robustness}
    \end{table}

    \begin{figure*}[t]
	
    \begin{subfigure}{\textwidth}

    \includegraphics[width=.9\textwidth]{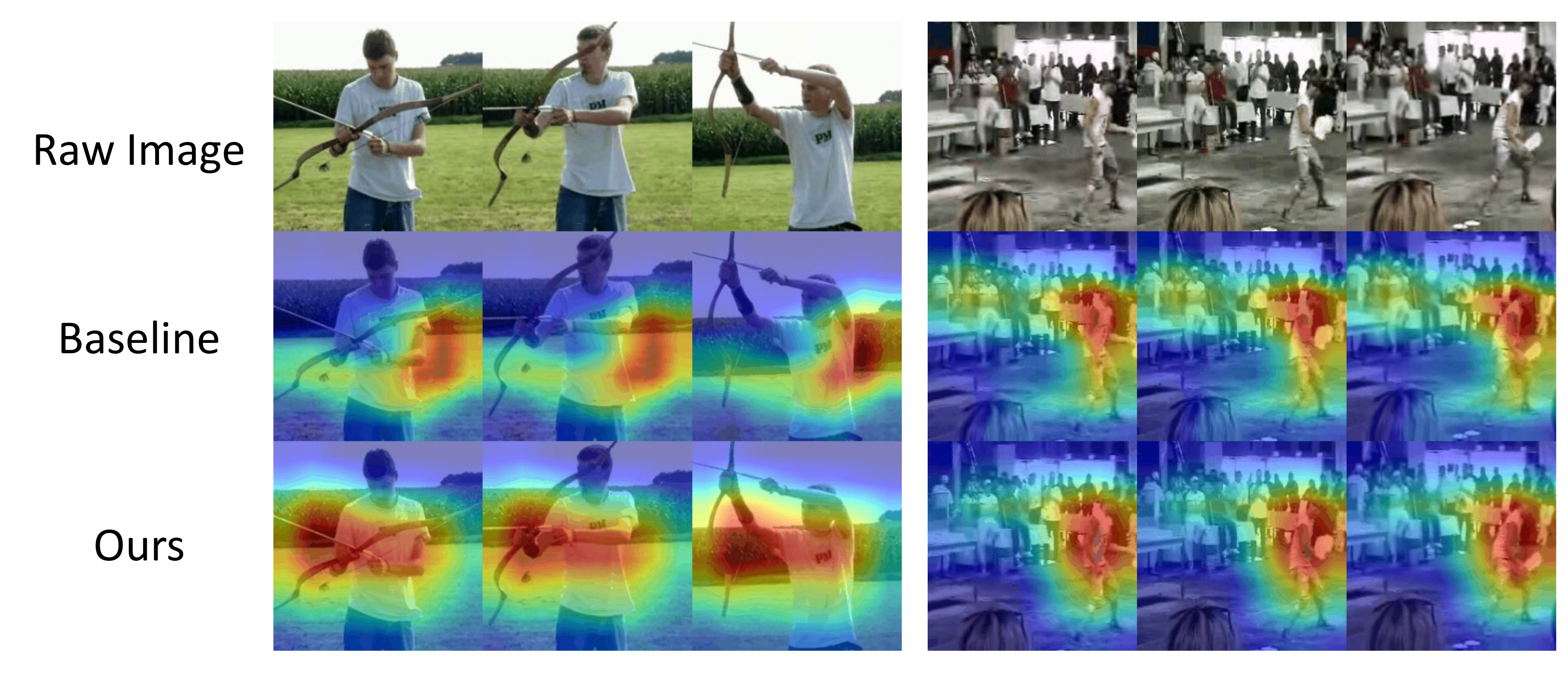}
    \end{subfigure}
    
    \begin{subfigure}{\textwidth}

    \includegraphics[width=.9\textwidth]{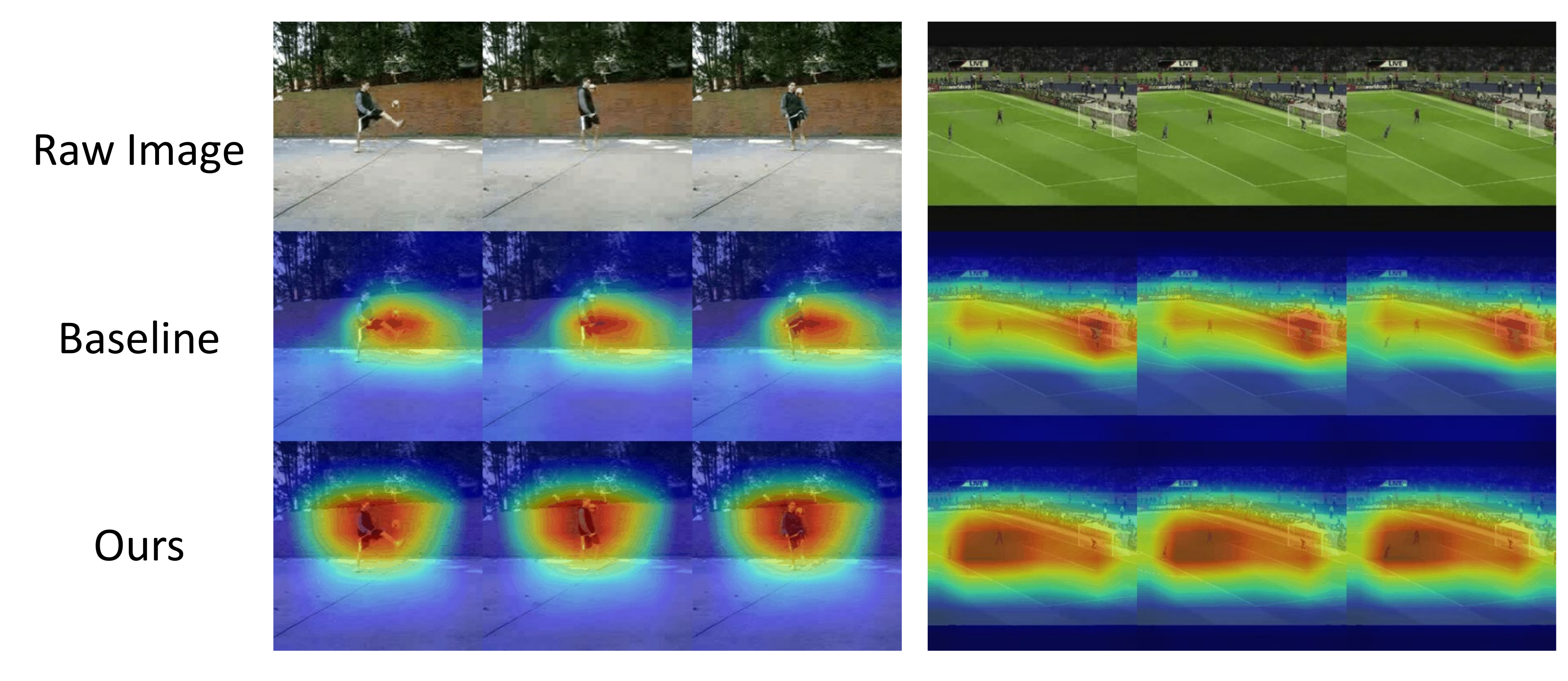}
    \end{subfigure}
    \begin{subfigure}{\textwidth}
    \includegraphics[width=.9\textwidth]{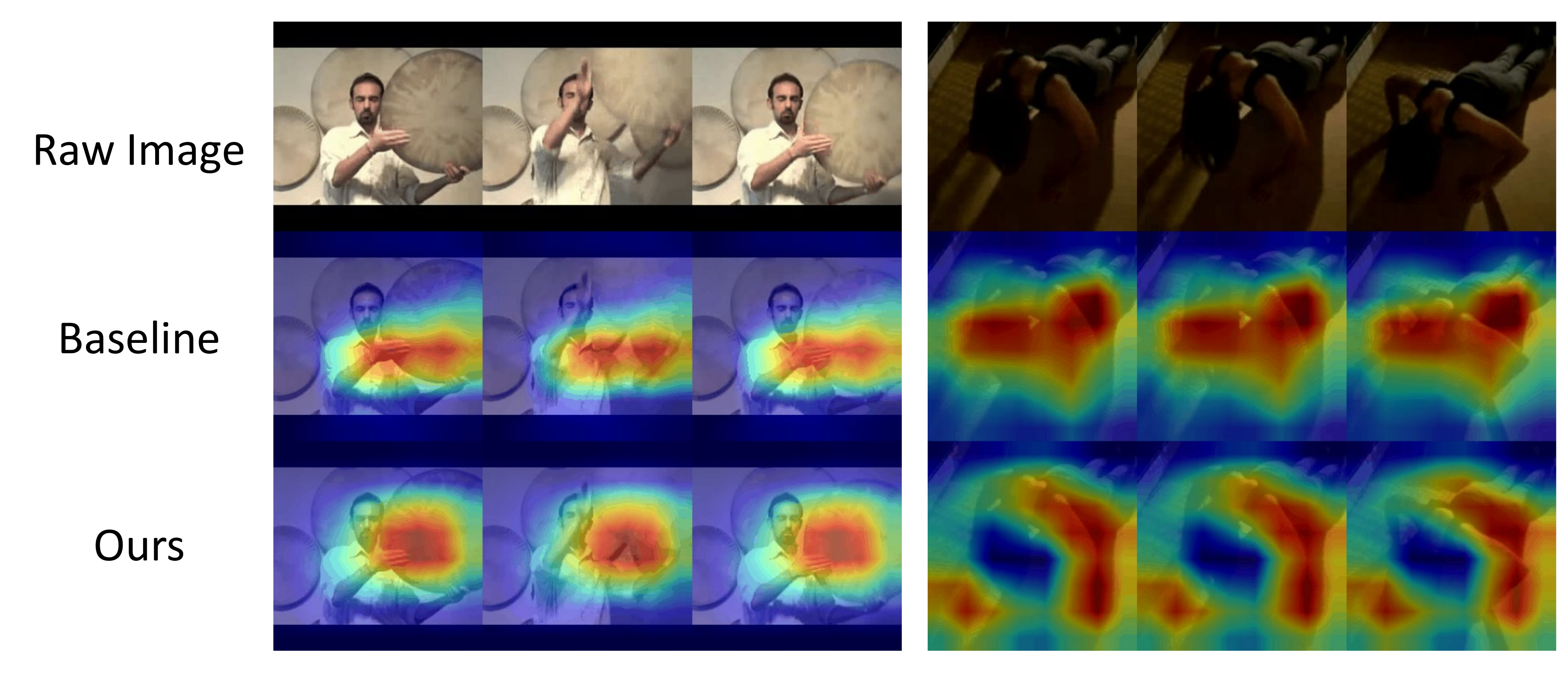}
    \end{subfigure}
    
    \captionsetup{width=.9\linewidth}
    \caption{\textbf{Grad-CAM visualization of the attention maps.} The videos are sampled from the validation set of UCF-101. The models are trained with 20\% labels (UCF101-20\%).}
    \label{fig:gradcam}
    \end{figure*}

    \begin{figure*}[t]\ContinuedFloat
    
    \begin{subfigure}{\textwidth}
    \includegraphics[width=.9\textwidth]{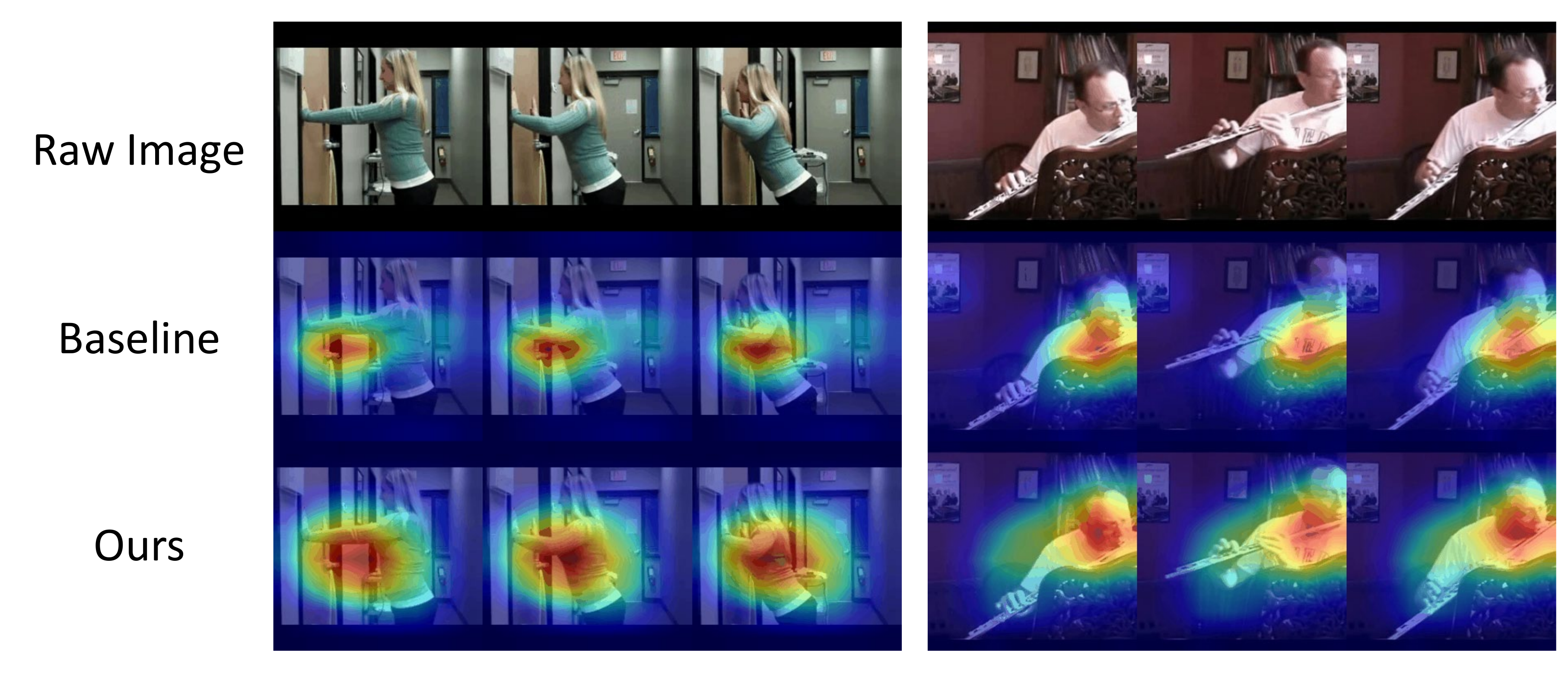}
    \end{subfigure}
    \captionsetup{width=.9\linewidth}
    \caption{\textbf{Visualization of the Grad-CAM attention maps.} The videos are sampled from the validation set of UCF-101. The models are trained with 20\% labels (UCF101-20\%).}
    \end{figure*}

    \begin{figure*}[htbp]
    \centering
    \includegraphics[width=0.9\textwidth]{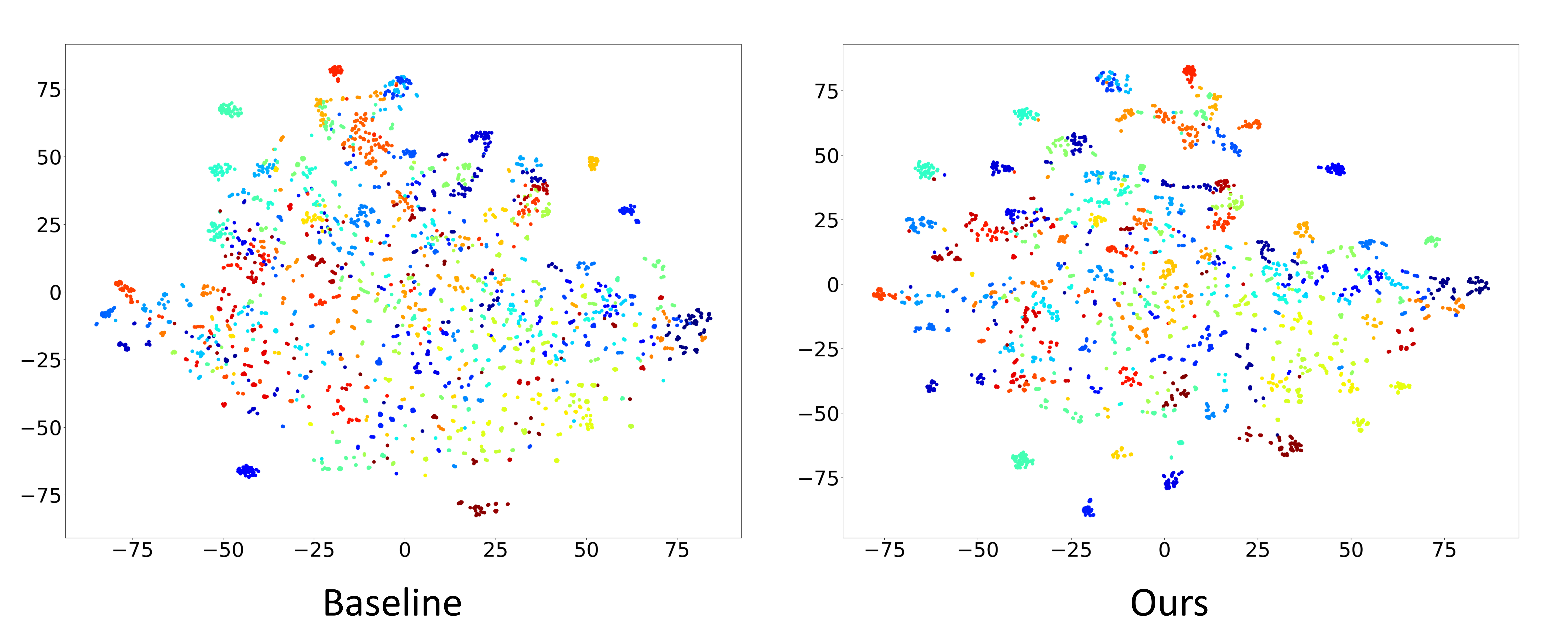}
    \vspace{-2mm}
    \captionsetup{width=.9\linewidth}
    \caption{\textbf{The comparison of t-SNE visualizations of the baseline and our method.} The visualized features are globally averaged features extracted by the backbone. All the videos of the validation set of UCF-101 are evaluated. The models are trained with 20\% labels (UCF101-20\%).}
    \label{fig:tsne}
    \vspace{-2mm}
    \end{figure*}
    
	\section{Grad-CAM Attention Maps} 
	\label{sec:gradcam}
	To better demonstrate that our method focuses more on the motion-related information, we visualize the attention maps with Grad-CAM~\cite{selvaraju2017grad} of multiple videos of UCF-101~\cite{soomro2012ucf101} validation set. As shown in \Cref{fig:gradcam}, the attention of our model is more reasonable and focuses more on the acting humans and moving objects.

\section{t-SNE Feature Visualization} 
\label{sec:tsne}
We also visualize the high-level features with t-SNE ~\cite{van2008visualizing} for showing a better latent representation space with our method. The visualization results covering the extracted features of the whole UCF-101~\cite{soomro2012ucf101} validation set are shown in \Cref{fig:tsne}. The features extracted with our method are more separable and easier to be classified in the latent representation space.

\section{Overfitting is Alleviated}
\label{sec:overfitting}

\Cref{tab:overfitting} presents a significant accuracy gap between the training and testing set, showing that FixMatch severely overfits to the training set. Our method effectively reduces the gap by imposing additional regularization on models with RGB as input.

\begin{table}[h]
\vspace{-2mm}
\tablestyle{3pt}{1.05}
\small
\begin{tabular}{l|ccc}
                   & \multicolumn{1}{l}{Training Acc.} & Testing Acc.             & Acc. Gap \\ \shline
Baseline-RGB       & 98.5                              & 54.1                     & 44.4     \\
Ours-RGB (Student) & 98.0                              & \multicolumn{1}{c}{76.1} & \textbf{22.0}     \\ 
\hline
Baseline-TG        & 97.4                              & 68.6                     & 28.8     \\
Ours-TG (Teacher)  & 96.5                              & 75.3                     & \textbf{21.2}    
\end{tabular}
\vspace{-3mm}
\caption{\textbf{Top-1 accuracy of the final models}. The models are trained on UCF-101 with 20\% labels.}
\vspace{-1mm}
\label{tab:overfitting}
\end{table}

\section{License of Used Assets}	

Kinetics-400~\cite{kay2017kinetics}: Creative Commons Attribution 4.0 International License; HMDB-51~\cite{kuehne2011hmdb}: Creative Commons Attribution 4.0 International License; UCF-101~\cite{soomro2012ucf101}: \href{https://www.crcv.ucf.edu/data/UCF101.php}{https://www.crcv.ucf.edu/data/UCF101.php}.

\end{document}